%% file: root.tex
\documentclass[letterpaper, 10 pt, conference]{ieeeconf}  %
\usepackage{caption}

\IEEEoverridecommandlockouts                              %

\input{macros}

\title{\LARGE \bf
\shortname: Adapting Whole-Body References for Force and Terrain Aware Humanoid Motion Tracking
}

\newcommand{\PublicProjectPage}{\url{https://multiplylabs.github.io/brace/}}

\author{ Sudarshan Harithas$^{1,2}$, Chen Yu$^1$ , Juan Borbon$^1$, Shubhankar Mondal$^3$, Winston Zha$^1$, \\ Srinath Sridhar$^{2}$,  Dingqi Zhang$^1$, Jiuguang Wang$^1$ \\ 
\PublicProjectPage
\thanks{$^1$Multiply Labs, USA: Work conducted while Sudarshan Harithas was at Multiply Labs.}
\thanks{$^2$Brown University, USA.}
\thanks{$^3$University of New Mexico, USA.}
}

\makeatletter
\let\@oldmaketitle\@maketitle
\renewcommand{\@maketitle}{%
  \@oldmaketitle
  \centering
  \centering
  \includegraphics[width=\textwidth]{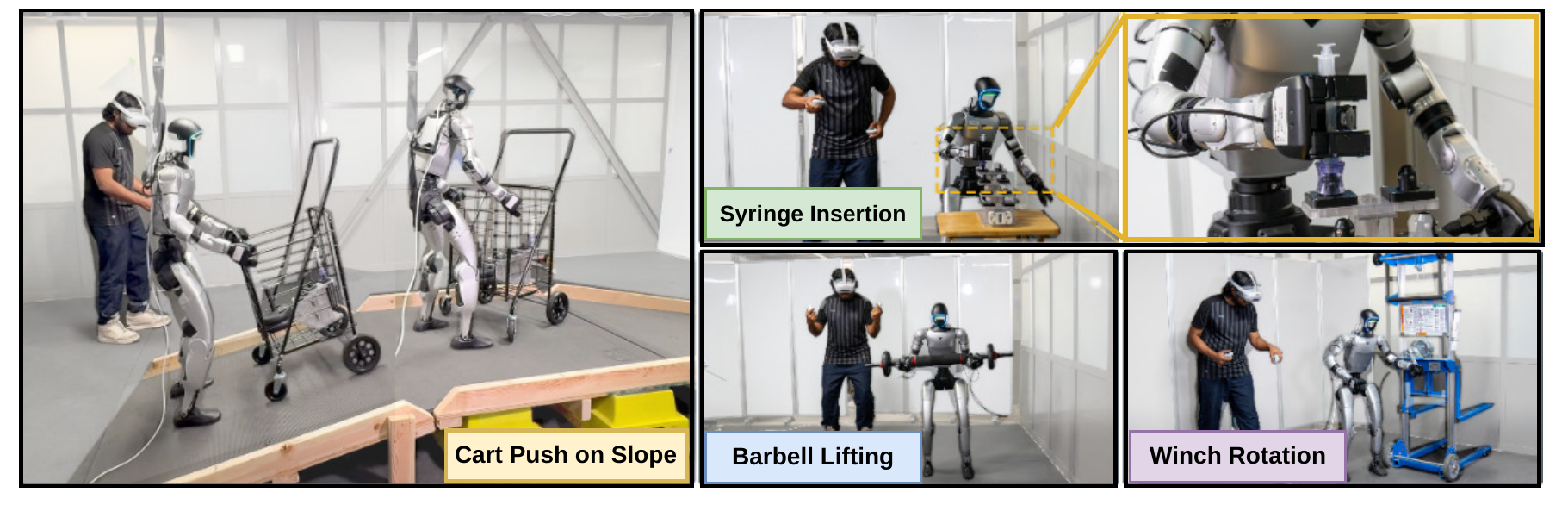}
  \protect\captionof{figure}{%
  \footnotesize{
      \textbf{\shortname{} is a Whole Body Tracker for Humanoid robots, which capable of exerting a commanded wrench value, this includes both linear force as required in the syringe insertion task, and angular torque which is required to rotate the winch. \shortname is terrain adaptive, allowing it to push a cart over the slope. Furthermore, the policy can lift heavy weights such as lifting a barbell.}
  }}
  \label{teaser}
  \vspace{-4mm}
}

\begin{document}
\bstctlcite{IEEEexample:BSTcontrol}

\maketitle
\setcounter{figure}{1}
\thispagestyle{empty}
\pagestyle{empty}

\begin{abstract}
\input{text/abstract}

\end{abstract}
\section{Introduction}
\input{text/introduction}
\section{Related Work}

\input{text/related_works}

\section{Method}

\input{text/method_propose}
\vspace{-2mm}
\section{Experiments \& Results}\label{sec:results}
\input{text/results}
\section{Conclusion}
\shortname{} adapts motion references to terrain and six-dimensional wrenches. One proprioceptive student tracks flat-ground references on slopes while exerting or compensating hand forces, without height maps or wrench measurements. It supports cross-site teleoperation and hardware demonstrations for VLA learning. At each setting's highest tested load or command, \shortname{} reduces global tracking error by 40\% and improves success by 18.25 percentage points on average (Sec.~\ref{sec:results}).
\bibliographystyle{IEEEtran}
\bibliography{references}

\end{document}

%% file: macros.tex
\usepackage[dvipsnames]{xcolor}
\usepackage{epsfig}
\usepackage{graphicx}
\usepackage{amsmath}
\usepackage{amssymb}
\usepackage{booktabs}
\usepackage{times}
\usepackage{mathtools}
\usepackage{float}
\usepackage{xspace}
\usepackage{caption}
\usepackage{subcaption}
\usepackage{tikz}
\usepackage{makecell}
\usepackage{multirow}
\usepackage{multicol}
\usepackage{booktabs}
\usepackage{pifont}
\usepackage{pgfplots}
\usepackage{nomencl}
\usepackage[english]{babel}
\usepackage[autostyle]{csquotes}
\let\labelindent\relax

\usepackage[inline]{enumitem}
\usepackage{stackengine} 
\usepackage{resizegather}
\usepackage{iftex}
\ifPDFTeX
  \usepackage[accsupp]{axessibility}
\fi
\usepackage{hyperref}

\usepackage[utf8]{inputenc} %
\usepackage[T1]{fontenc}    %
\usepackage{hyperref}       %
\usepackage{url}            %
\usepackage{booktabs}       %
\usepackage{amsfonts}       %
\usepackage{nicefrac}       %
\usepackage{microtype}      %
\usepackage{graphicx}
\usepackage{caption}
\usepackage{amsmath}
\usepackage{amssymb}
\usepackage{mathtools}
\usepackage{wrapfig}
\usepackage{tabularx}
\usepackage{fontawesome5}
\usepackage[utf8]{inputenc} %
\usepackage[T1]{fontenc}    %
\usepackage{hyperref}       %
\usepackage{url}            %
\usepackage{booktabs}       %
\usepackage{amsfonts}       %
\usepackage{nicefrac}       %
\usepackage{microtype}      %
\usepackage{multicol}
\usepackage{multirow}
\usepackage{adjustbox}
\usepackage{wrapfig}
\usepackage{changes}
\usepackage{xspace}
\usepackage{float}
\usepackage{cleveref}
\usepackage{colortbl} %
\usepackage{cite}

\newcommand{\eg}[1]{\emph{e.g.,}\xspace}

\newcommand{\filluptopage}[1]{%
  \clearpage
  \loop\ifnum\value{page}<#1\relax
    \null\clearpage
  \repeat
  \loop\ifnum\value{page}=#1\relax
    \null\clearpage
  \repeat
}

\makeatletter
\def\blfootnote{\xdef\@thefnmark{}\@footnotetext}
\makeatother

\newcommand{\circlednumber}[1]{\raisebox{.5pt}{\textcircled{\raisebox{-.9pt} {#1}}}}

\newcommand{\shortname}{BRACE\xspace}

\newcommand{\wbt}{WBT\xspace}

\newcommand{\rom}[1]{\uppercase\expandafter{\romannumeral #1\relax}}

\usepackage{tcolorbox}

\newtcbox{\surface}{on line,
  colback=green!20,      %
  colframe=green!50!black, %
  boxrule=1pt,           %
  arc=1pt,               %
  left=1pt,right=1pt,top=1pt,bottom=1pt  %
}

\newtcbox{\reorientation}{on line,
  colback=cyan!20,      %
  colframe=cyan!50!black, %
  boxrule=1pt,           %
  arc=1pt,               %
  left=1pt,right=1pt,top=1pt,bottom=1pt  %
}

\newtcbox{\rigid}{on line,
  colback=purple!20,      %
  colframe=purple!60!black, %
  boxrule=1pt,           %
  arc=1pt,               %
  left=1pt,right=1pt,top=1pt,bottom=1pt  %
}

\newtcbox{\articulated}{on line,
  colback=yellow!80,      %
  colframe=yellow!50!black, %
  boxrule=1pt,           %
  arc=1pt,               %
  left=1pt,right=1pt,top=1pt,bottom=1pt  %
}

\newtcbox{\interactionfree}{on line,
  colback=pink!20,      %
  colframe=pink!50!black, %
  boxrule=1pt,           %
  arc=1pt,               %
  left=1pt,right=1pt,top=1pt,bottom=1pt  %
}

\newtcbox{\bimanual}{on line,
  colback=yellow!20,      %
  colframe=yellow!50!black, %
  boxrule=1pt,           %
  arc=1pt,               %
  left=1pt,right=1pt,top=1pt,bottom=1pt  %
}
\definecolor{magenta}{HTML}{C71585}  %

%% file: text/abstract.tex
Whole-body tracking has become the interface through which operators drive humanoid robots, yet the references it consumes are recorded on level ground and carrying nothing, so the tracker is aware of neither the forces the robot must exchange with objects nor the terrain it must stand on. 
Existing controllers address one side of this gap: force-capable policies command an end-effector force but prescribe no whole-body pose, while terrain-adaptive trackers treat loads as disturbances to reject rather than wrenches to command. 
We present \shortname, a whole-body tracker that exerts and compensates commanded hand forces from diverse poses while following a flat-ground reference on sloped terrain. Rather than leaving the tracker to absorb the load and the slope, \shortname folds both into the reference it follows: terrain conformance lifts footholds and root onto the local surface, and a wrench transformation resolves the hand displacement that produces a force jointly with the center-of-mass and center-of-pressure shifts it induces, bounded by teacher-specific arm-effort limits. 
Separate exertion and compensation teachers are distilled by DAgger into one flow-matching student that runs on proprioception alone, without a height map or measured wrench, so a teleoperator (even in remote locations) can supply a flat-ground trajectory and the robot resolves slope and load onboard. 
We extensive experiments on the Unitree-G1 to demonstrate the ability of \shortname to exert, compensate force and handle diverse terrain in simulation and real robot experiments.

%% file: text/introduction.tex
Whole-body tracking has become the standard route from human motion to humanoid
motion: a trajectory retargeted from a demonstrator is followed by a controller that keeps the
robot upright while it reproduces the
reference~\cite{sonic,twist2,any2track,beyondmimic,xie2025kungfubot,omnih2o,exbody2}. The
approach has scaled because its data has. Motion capture~\cite{AMASS} and in-the-wild
video~\cite{uh1} supply reference trajectories in quantity, and recent systems use the
resulting trackers as a teleoperation interface, collecting demonstrations for downstream
policies~\cite{sonic,twist2,omnih2o}.

What those corpora do not contain is force. A demonstrator is recorded on level ground,
carrying nothing, and the record preserves the path the body took but not the loads it bore or
the reactions it absorbed. For locomotion and gesture little is lost. For the work humanoids
are meant to do, what is lost is the specification: moving a payload, pulling a
door against its closer, or steadying a workpiece while turning a tool is defined by the wrench
$\mathbf{w} = [\,F,\, \tau\,]$ the robot exchanges with its environment at least as much as by
the trajectory it follows. Terrain is missing for the same reason, and under teleoperation the
two omissions compound: the operator stands unloaded in a capture volume on level ground while
the robot walks a ramp with a box in its hands. Pushing a cart up a grade is the simple case,
where the slope changes both the force the robot must supply and the lean it must adopt to
supply it, and the retargeted motion shows neither.

Existing controllers address one or the other. Force-capable legged systems accept an
end-effector force alongside a base velocity~\cite{unifp,falcon,hafo,kinaware}, or accept
impedance parameters in place of a force~\cite{facet}; since none prescribes a dense whole-body
pose, the question of producing a force while holding a posture does not arise.
Terrain-adaptive controllers infer ground geometry from
proprioception~\cite{rma,loquercio2023learning,any2track}, but treat loads and pushes as
disturbances to reject rather than wrenches to command.

Combining the two inside a whole-body reference is difficult because each signal contradicts
that reference. A commanded wrench is conventionally realized by displacing the end
effector~\cite{unifp,chip}; displace it without adjusting the rest of the body and the robot
loses balance once the reaction arrives, while the pose the tracker is asked to hold diverges
from the pose that would produce the force. The operator's support surface is likewise not the
robot's, leaving reference footholds and root height systematically wrong, and no dataset pairs
retargeted humanoid motion with the terrain it was executed on, so the correction cannot be
learned directly.

\shortname (Body Reference Adaptation for Contact and Elevation) resolves this conflict at its source by correcting the reference rather than the controller. A flat-ground reference is adapted analytically to the terrain the robot stands on and to the wrench it is asked to exert, so that tracking the adapted reference already yields the footholds, hand displacement, and bracing the task requires. Two privileged teachers learn on these references, one that exerts a commanded wrench and one that holds the reference against an external one, and both are distilled into a single flow-based student that runs from proprioception and operator commands alone, inferring terrain and load on the fly. \shortname focuses on proprioception centric WBC and deals with non-flat terrains that can be handled with proprioception~\cite{rma}, which include slopes, rough surface inconsistencies, and small platforms.  
To our knowledge, \shortname is the first whole-body tracker that exerts or rejects a wrench and adapts to terrain at the same time, enabling tasks such as inserting a syringe, turning a winch, lifting a barbell, and pushing a cart up a slope (Fig.~\ref{teaser}). 
Our results (Sec.~\ref{sec:results}) indicate that \shortname reduces global drift at an average of $40$ percentage points and results in an overall average improvement in success rate by $18.25$ percentage points. 
We further deploy the \shortname policy on the Unitree-G1 to demonstrate its operation in the real world. 

%% file: text/related_works.tex
\subsection{Whole-Body Tracking on Terrain and Under Load}
Reinforcement learning has been highly successful for training motion imitation models in character animation~\cite{pulse,maskedmimic,peng2018deepmimic} and humanoid control~\cite{sonic,asap,twist2,beyondmimic,omnih2o,exbody2}. Recent work has pushed whole-body tracking (\wbt) toward real-world robustness~\cite{any2track,heft,asap}. 
Any2Track~\cite{any2track} separates tracking from dynamics adaptation to handle sim-to-real mismatch and diverse terrain, while HEFT~\cite{heft} focuses on tracking under hand-carried payloads. 
These advances improve motion tracking under environmental disturbances, but leave open the joint problem of tracking whole-body motion and exerting commanded wrenches across varied terrain.

In particular, HEFT handles payloads as quasi-static gravitational loads, and Any2Track treats external pushes as unmodeled perturbations; neither exposes a commanded interaction wrench to the operator. 
Other terrain-aware approaches generate motion references online from perception~\cite{wbloco2026,perceptivebfm}, coupling the reference to the robot's local terrain. 
\shortname{} instead accepts a flat-ground reference and internally adapts tracking to the terrain and interaction task. 
This allows the operator to specify the intended motion and wrench while the controller reconciles them with the robot's support surface and loading conditions. 
Such a separation supports applications including remote teleoperation, where the operator and robot need not share the same terrain or physical load (Sec.~\ref{sec:applications}).

\subsection{Force Control in Locomotion}
Force exertion and compensation are two key paradigms in force control that in focus of \shortname. 
For force exertion, UNIFP~\cite{unifp} commands an end-effector force alongside a base velocity, while FACET~\cite{facet} commands impedance rather than force, imitating a virtual spring--damper at the CoM. 
For force compensation, which asks the robot to hold a pose against external forces, decoupled frameworks such as FALCON~\cite{falcon}, HAFO~\cite{hafo}, and KinAware~\cite{kinaware} pair a lower-body agent for standing with an upper-body agent for end-effector tracking under force curricula. 

What these interfaces share is the absence of a dense whole-body pose target: the operator supplies a root velocity or an end-effector goal, and the remaining degrees of freedom are resolved by the controller rather than prescribed. \shortname{} instead holds a full whole-body reference against a commanded wrench, so the wrench must be reconciled with the reference rather than absorbed by unconstrained degrees of freedom. Moreover, each of these methods addresses a single paradigm, whereas \shortname{} must distinguish producing a commanded wrench from holding a motion against an external load, and switch between them on command.

\subsection{Behavior Models and Multi-Teacher Distillation}
Generalist controllers consolidate diverse skills and command interfaces into a shared policy. In character animation, PULSE~\cite{pulse} learns a variational motion prior that can be reused for downstream tasks, while MaskedMimic~\cite{maskedmimic} supports diverse control inputs through masked motion conditioning. Related approaches extend generalist control to humanoid robots: HOVER~\cite{hover} distills multiple control modes into one controller, and BFM-Zero~\cite{bfmzero} learns a shared latent representation of motions, goals, and rewards. BeyondMimic~\cite{beyondmimic} uses guided latent diffusion to compose motions and specify tasks, while OmniXtreme~\cite{omnixtreme} distills motion-specific experts into a flow-matching policy through DAgger.

Building on these approaches, \shortname{} addresses the joint demands of whole-body motion tracking, wrench interaction, and terrain adaptation. The controller must distinguish producing a commanded wrench from preserving a motion against an external load, even in similar body configurations. We distill separate privileged exertion and compensation teachers into one student through DAgger~\cite{dagger}. A motion goal, wrench command, and explicit interaction mode specify the desired response. This formulation extends multi-teacher distillation to complementary interaction behaviors while retaining whole-body motion tracking across terrain and loading conditions.

%% file: text/method_propose.tex
\begin{figure*}[t]
    \captionsetup{font=footnotesize,skip=4pt}
    \centering
    \includegraphics[width=\textwidth]{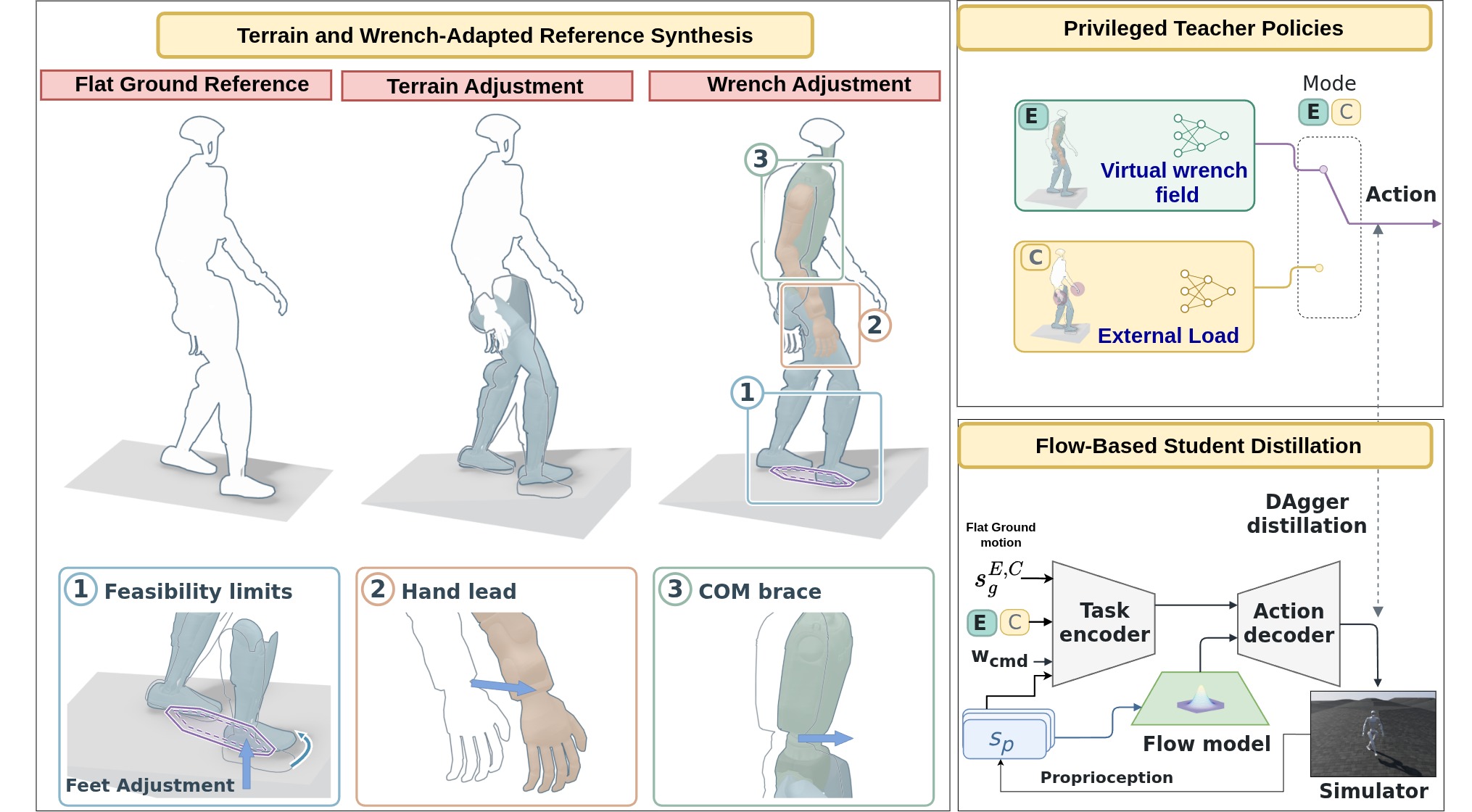}
    \caption{\textbf{\shortname overview.} \emph{Left:} terrain- and wrench-adapted reference synthesis (Sec.~\ref{sec:reference}). The flat-ground reference $x_{\mathrm{ref}}$ is adjusted to the terrain (given by the heightmap $\mathcal{H}$) through foot alignment, giving $x_{\mathrm{ref}}^T$, and to the commanded wrench through \circlednumber{1} feasibility limits, \circlednumber{2} hand lead, and \circlednumber{3} COM brace, giving $x_E^\star$. \emph{Top right:} privileged teacher policies (Sec.~\ref{sec:teachers}). Teacher-E learns force exertion under a virtual wrench field while tracking $x_E^\star$; Teacher-C learns force compensation under an external wrench while tracking $x_C^\star = x_{\mathrm{ref}}^T$. \emph{Bottom right:} flow-based student distillation (Sec.~\ref{sec:student}). DAgger distills both teachers, selected by the mode flag $\mu$, into a single student: a task encoder consumes the flat-ground goals $s_g^{E,C}$, the mode, and the wrench command $w_{\mathrm{cmd}}$, a flow-matching base models the action distribution from the proprioceptive history $s_p$, and an action decoder fuses both into the executed action.}
    \vspace{-6mm}
    \label{fig:method}
\end{figure*}

\shortname enables humanoid whole-body motion tracking under force exertion and force compensation on uneven terrain. The framework operates given two operator inputs: a flat-ground kinematic reference trajectory $x_{\mathrm{ref}}$ retargeted from human motion capture~\cite{sonic,twist2,heft,any2track}, and a set of commanded hand wrenches $w_{\mathrm{cmd}} = \{w_h\}_h$, where $w_h = [F_h; M_h] \in \mathbb{R}^6$ specifies the force $F_h$ and moment $M_h$ for hand $h$.

The kinematic reference $x_{\mathrm{ref}}$ conflicts with both the terrain geometry $\mathcal{H}$ on which the robot operates and the commanded wrenches $w_{\mathrm{cmd}}$. Kinematically, $x_{\mathrm{ref}}$ assumes a flat floor, causing foothold and root errors on uneven terrain. Dynamically, exerting or resisting a wrench requires whole-body bracing and contact displacement, neither of which exists in unloaded human recordings. Training an RL policy to discover these corrections via auxiliary rewards creates a signal conflict, as trajectory tracking rewards penalize the necessary pose deviations. \shortname eliminates this conflict at its source by analytically transforming $x_{\mathrm{ref}}$ according to $\mathcal{H}$ and $w_{\mathrm{cmd}}$ prior to policy learning, reducing the learning problem to tracking geometrically adapted, effort-limited targets.

Fig.~\ref{fig:method} summarizes the framework in three parts, which the following subsections mirror. Reference synthesis (left; Sec.~\ref{sec:reference}) analytically adapts $x_{\mathrm{ref}}$ to the terrain and to the commanded wrench, producing one tracking target per teacher. Privileged teacher training (top right; Sec.~\ref{sec:teachers}) trains Teacher-E for force exertion and Teacher-C for force compensation on these targets with a shared tracking architecture. Student distillation (bottom right; Sec.~\ref{sec:student}) distills both teachers into a single flow-based student that operates from onboard observations alone.

\subsection{Terrain- and Wrench-Adapted Reference Synthesis}
\label{sec:reference}

We start with a dataset of $G1$ reference trajectories collected on flat-land without any terrain information, the trajectory is represented as $x_{ref}$. 
Now given a terrain represented by the height map ($\mathcal{H}$), the function $\mathcal{T(.)}$ given in Eq.~\ref{eq:teacher_targets} transforms the reference to map to terrain leading to the trajectory $x_{ref}^{T}$. 
The \emph{Teacher-C}, the compensation teacher is trained on $x_{ref}^{T}$. 
The exertion teacher takes an additional input $w_{cmd}$, and the function $\mathcal{F(.)}$ is used to perform the wrench adaptation. 

\begin{equation}
\begin{aligned}
x_{\mathrm{ref}}^T &= \mathcal{T}(x_{\mathrm{ref}}, \mathcal{H}), \\
x_E^\star &= \mathcal{F}(x_{\mathrm{ref}}^T, w_{\mathrm{cmd}}), \\
x_C^\star &= x_{\mathrm{ref}}^T.
\end{aligned}
\label{eq:teacher_targets}
\end{equation}

The left panel of Fig.~\ref{fig:method} illustrates the terrain adjustment and the three components \circlednumber{1}--\circlednumber{3} of the wrench adjustment, which we detail below.

\paragraph{Terrain Adjustment $\mathcal{T}$}
Kinematic retargeting ignores local surface height, leading to foothold errors on slopes and steps. For each foot contact target $(x_f, y_f, z_f^{\mathrm{ref}})$, $\mathcal{T}$ preserves horizontal placement and lifts the elevation to $z_f^T = \mathcal{H}(x_f, y_f) + z_f^{\mathrm{ref}}$, which maintains swing clearance relative to the local ground height. Root elevation follows the terrain beneath the base, and the local surface normal rotates the foot orientation targets while the root orientation stays aligned with the locomotion direction. %
A whole-body inverse kinematics (IK) step reconciles foot and root adjustments with the upper-body pose, producing the terrain-adjusted reference $x_{\mathrm{ref}}^T$.

\paragraph{Wrench Adjustment $\mathcal{F}$ (\circlednumber{1}--\circlednumber{3})}
Generating external forces requires end-effector displacement into the contact and a shift of the center of mass (COM) to counteract the reaction. Wrench adjustment $\mathcal{F}$ first caps the commanded wrench $w_{\mathrm{cmd}}$ to a feasible wrench $w^\star = [F^\star; M^\star]$ and then converts $w^\star$ into coordinated hand leads and COM shifts, the three callouts \circlednumber{1}--\circlednumber{3} in Fig.~\ref{fig:method}.

\begin{itemize}[leftmargin=*] %
\item \textit{Feasibility limits (\circlednumber{1}):} Both teachers must decide which wrenches the arms can support: Teacher-E to cap what it asks the robot to exert, Teacher-C to cap what it applies as a disturbance. We use one limit for both. Force and moment consume the same arm-joint effort, so the two are constrained jointly rather than separately. Omitting the hand index and writing a single hand's force and moment as $F, M$, the translational and rotational arm Jacobians $J_p, J_r$ map the wrench to induced joint torques, which a teacher-specific margin $0 < \gamma < 1$ holds inside a joint-torque budget $\tau_i^{\mathrm{budget}}$:
\begin{equation}
\bigl| \bigl( J_p^\top F + J_r^\top M \bigr)_i \bigr| \le \gamma \, \tau_i^{\mathrm{budget}}.
\label{eq:effort_budget}
\end{equation}
The budget itself differs by teacher. Teacher-E uses the actuator limit $\tau_i^{\mathrm{budget}} = \tau_i^{\max}$ evaluated at the terrain-adjusted reference, whereas Teacher-C uses a floored residual-effort proxy evaluated at the current robot configuration (Sec.~\ref{sec:teachers}). For Teacher-E, this constraint is one of the caps that turn $w_{\mathrm{cmd}}$ into $w^\star$: unbounded commands risk requesting joint torques that exceed motor limits or destabilize contact, so hand-excursion bounds and Eq.~\eqref{eq:effort_budget} constrain $w^\star$, while the contact-weighted foot support polygon and its eroded inner margin bound center-of-pressure travel and the COM brace. A shared support-based scale limits both hands' force and moment commands, smoothed and lower-bounded across gait transitions.
\item \textit{Hand lead (\circlednumber{2}):} Task-space force generation is modeled as spring--damper compliance along the force direction $\hat{n} = F^\star / \|F^\star\|$~\cite{unifp}. Given diagonal joint stiffness $K_q$ and damping $D_q$, the directional stiffness $k_F$ and hand displacement $\delta_p$ satisfy
\begin{equation}
k_F^{-1} = \hat{n}^\top J_p K_q^{-1} J_p^\top \hat{n}, \qquad
d_F \dot{\delta}_p + k_F \delta_p = \|F^\star\|,
\label{eq:reference_dynamics}
\end{equation}
with the directional damping $d_F$ defined analogously from $D_q$. Deriving $k_F$ from the joint gains matches the lead to the compliance of the PD-controlled arm, so reaching the displaced target produces $F^\star$. Rotational leads driven by the rotational stiffness $k_M$ and moment direction $\hat{m} = M^\star / \|M^\star\|$ follow the same formulation. Directions are defined only for nonzero commands: commands below $1\,\mathrm{N}$ or $0.3\,\mathrm{N}\!\cdot\!\mathrm{m}$ are zeroed, so the corresponding lead vanishes and the directional stiffness and damping revert to their default constants.
\item \textit{COM brace (\circlednumber{3}):} Sustained hand wrenches generate overturning moments about the feet that must be counteracted by leaning the torso. Countering the reaction wrench $-w^\star$ requires a horizontal COM shift $\Delta c$:
\begin{equation}
\Delta c = \frac{\hat{z} \times \sum_h \bigl[ (p_h - c_{\mathrm{cop}}) \times (-F_h^\star) - M_h^\star \bigr]}{m g},
\label{eq:support_brace}
\end{equation}
where the hand index is restored for the two-hand sum, $p_h$ is the contacting hand position, $c_{\mathrm{cop}}$ is the contact-weighted foot centroid approximating the center of pressure, $m$ is the robot mass, $g$ is gravitational acceleration, and $\hat{z}$ is the vertical unit vector. Center-of-pressure travel within the support polygon absorbs part of $\Delta c$; the bounded remainder forms the brace, half of which the hand targets inherit.
\end{itemize}
Whole-body IK reconciles the displaced hand targets, braced COM position, and terrain-adjusted foot positions into a kinematically valid adapted target $x_E^\star$.

\subsection{Privileged Teacher Policies}
\label{sec:teachers}

Both teachers track the adapted targets with the same architecture (Fig.~\ref{fig:method}, top right); the arm-effort constraint of Eq.~\eqref{eq:effort_budget} bounds what Teacher-E exerts and what Teacher-C resists.

\paragraph{Shared Architecture}
Following goal-space redirection~\cite{sonic}, the actor observes a goal derived from the flat-ground reference ( represented as $s_g^E$ for Teacher-E and $s_g^C$ for Teacher-C) rather than the adapted target $x^\star$, on which the reward is evaluated. Its remaining inputs are proprioception (joint positions and velocities, base angular velocity, and gravity direction), a short observation history, and a privileged latent. The privileged encoder compresses the terrain height scan, the root height above the terrain, and teacher-specific channels (the surface normal for Teacher-E, the applied-wrench history for Teacher-C) into this latent.

Two heads share the policy trunk. The action head commands joint position targets for the $29$ actuated joints, and an auxiliary reconstruction head predicts $x^\star$, which forces the trunk to represent the full-body pose corrections. Both teachers are trained with PPO and asymmetric critics that additionally observe the full simulator state; only the action head is retained after training.

\paragraph{Teacher-E: Force Exertion}\label{sec:force_exert}
Teacher-E observes the terrain-invariant goal $s_g^E$: the flat-ground reference offset by the hand leads and COM brace of Sec.~\ref{sec:reference}, with the absolute root elevation removed. The wrench command reaches the policy only through this displacement. Tracking alone does not exert anything, since in an empty simulator the hand reaches its displaced target without producing force. A \textit{virtual wrench field}~\cite{softmimic} anchored at the terrain-adjusted reference hand therefore supplies the reaction. For a hand translation $p$ along $\hat{n}$ and twist $\theta$ about $\hat{m}$, both measured relative to the torso pose,
\begin{equation}
f_{\mathrm{meas}} = k_F (\bar{p} + b/2) + d_F \dot{p}, \qquad
m_{\mathrm{meas}} = k_M \theta + d_M \dot{\theta},
\label{eq:virtual_field}
\end{equation}
where $\bar{p}$ is $p$ smoothly bounded to the allowed excursion and $b$ is the planned brace along $\hat{n}$. The term $b/2$ credits the half of the brace not inherited by the hand targets to the spring force, so leaning into the contact counts toward exertion. The simulator applies the reaction $-w_{\mathrm{meas}} = -[f_{\mathrm{meas}}\hat{n}; m_{\mathrm{meas}}\hat{m}]$ at the hand. Rewards score tracking against $x_E^\star$, normalized wrench error, and motion smoothness. The wrench error is measured against the feasible target $w^\star$ rather than $w_{\mathrm{cmd}}$, so the policy is judged by what the terrain and joint budget permit.
The overall rewards for the exertion teachers include tracking and wrench exertion rewards.$r_{exert} = r_{tracking} + r_{wrench}$, where $r_{wrench}$ measures the difference between the commanded wrench ($w_{cmd}$) and exerted wrench in the simulator.

\paragraph{Teacher-C: Force Compensation}\label{sec:force_comp}
Teacher-C tracks $x_C^\star = x_{\mathrm{ref}}^T$ while the simulator applies an external wrench $w_{\mathrm{ext}}$ at the hands (Fig.~\ref{fig:method}, top right). Resisting such wrenches requires holding the reference posture without collapsing under load or over-correcting when the load vanishes. Following the torque-aware curriculum of FALCON~\cite{falcon}, loads are bounded by Eq.~\eqref{eq:effort_budget} evaluated at the current robot configuration, with the floored residual-effort budget
\begin{equation}
\tau_i^{\mathrm{budget}} = \max \bigl( \tau_i^{\max} - |\tau_i|,\, \epsilon \tau_i^{\max} \bigr),
\label{eq:residual_budget}
\end{equation}
where $\tau_i$ is the torque already spent on gravity and tracking and $\epsilon > 0$ preserves a minimum range (pink region in Fig.~\ref{fig:method}, top right). A random application offset $r_h$ in the hand frame adds the moment $r_h \times F_h$, so force and moment are constrained jointly. Load profiles sweep force and moment through randomized triangular pulses separated by unloaded intervals, which expose the policy to load release, and an adaptive curriculum raises each environment's load scale after sustained success and lowers it after falls. The privileged encoder receives a short history of the applied wrench, which lets the teacher brace before a load peaks. Teacher-C receives tracking and regularization rewards only; compensation emerges from tracking error minimization under load.

\subsection{Flow-Based Student Distillation}
\label{sec:student}

\paragraph{Distillation}
The teachers depend on privileged terrain and wrench signals that the robot lacks. We therefore distill both into a single deployable student with DAgger~\cite{dagger}. A mode flag $\mu \in \{\mathrm{E}, \mathrm{C}\}$ selects the active teacher during training rollouts (the E/C switch in Fig.~\ref{fig:method}). Rollouts include regime-switching transitions (ramping the active load to zero, switching the mode flag, and ramping the incoming wrench), so that mode switches mid-motion are covered during training. 
Observe that the student only consumes the flat terrain motion $s_{g}^{E,C}$ which is available at inference and can be provided by a teleoperator. 

\paragraph{Student Architecture}
Exerting a wrench and resisting an external load require different joint actions from nearly identical proprioceptive states. The goal-free base is conditioned only on proprioceptive history, so its conditional action distribution is multimodal; flow matching represents that distribution rather than averaging it. The student employs three blocks (Fig.~\ref{fig:method}, bottom right):
\begin{enumerate}[leftmargin=*]
\item \textit{Flow-matching base:} A flow-matching network~\cite{omnixtreme,beyondmimic} maps Gaussian noise to a base joint position action conditioned on a 25-frame proprioceptive history $s_p$. Operating without goals or mode flags, it models the distribution of teacher actions given proprioception.
\item \textit{Task encoder:} A deterministic encoder maps proprioception and its history, together with the flat-ground goals $s_g^{E,C}$, the mode flag $\mu$, and the wrench command $w_{\mathrm{cmd}}$, into a compact task latent.
\item \textit{Action decoder:} Combines the base action and the task latent into the executed joint commands.
\end{enumerate}
The student is trained with behavior cloning on the selected teacher's actions, a flow-matching loss on the base, an exertion loss that weights action errors by their sensitivity to hand force, and the estimation regression below. %

\paragraph{Terrain and Load Estimation}
The student policy is directly deployed on the real robot, onboard sensors observe neither the terrain geometry nor the external wrench, so the student must infer both from proprioceptive dynamics. An auxiliary estimation head regresses the privileged simulator states (local hand positions, root velocity, external wrench, surface normal, and root height above the terrain) from the task latent during training. The head is discarded at export; the deployed student observes only proprioception, goals, mode flag, and wrench command.

\subsection{Dataset, Implementation and real robot deployment}
\label{sec:impl}

All policies command PD position targets for the 29-DoF Unitree G1 at $50\,\mathrm{Hz}$; the flow base is integrated with four Euler steps per action. %
Teachers use an 8-frame observation history and the student a 25-frame history. The height scan is a $16 \times 16$ grid covering $2.0\,\mathrm{m} \times 2.0\,\mathrm{m}$ in the base frame, and the wrench history spans 3 frames. %
Both teachers use $\gamma = 0.7$ in the shared arm-effort limit. In $\mathcal{F}$, the hand-excursion budget is $0.15\,\mathrm{m}$ and the support polygon is eroded by $0.02\,\mathrm{m}$; the directional stiffnesses $k_F, k_M$ are lower-bounded for numerical stability, and the support-polygon scale is low-pass filtered to remain smooth across double-to-single support transitions. Our compensation teacher is trained to compensate upto $20N$ force, and the exertion teacher is trained upto $9N$, the torque limits for both teachers is upto $4Nm$. These settings are done to be within the phsycial limits of the G1 robot's arms~\cite{unitree_g1_specs,unitree_ros_g1_urdf}.

The teachers and the student are trained on 8x NVIDIA RTX Pro 6000 GPUs each with a $2048$  environments and $8192$ batch size per GPU.
The RL teacher and the student is trained in IsaacLab~\cite{mittal2025isaaclab} which runs on IsaacSim. 
Specifically we use the Protomotions~\cite{ProtoMotions} which provides the learning  framework for learning, and multi-simulator inference. 
Following Protomotions~\cite{ProtoMotions} our real-robot deployment infrastructure for the flow policy is built on RoboJudo~\cite{RoboJuDo}.
For training we use reference trajectories from the Bones-Seed~\cite{bones_seed_2026}, we do not use the complete dataset, instead we use text semantics to select locomotion and object manipulation data from the complete dataset. 
This leads to around $65k$ reference trajectories for training.

%% file: text/results.tex
\begin{figure*}[t]
    \captionsetup{font=footnotesize,skip=4pt}
    \centering
    \includegraphics[width=\textwidth]{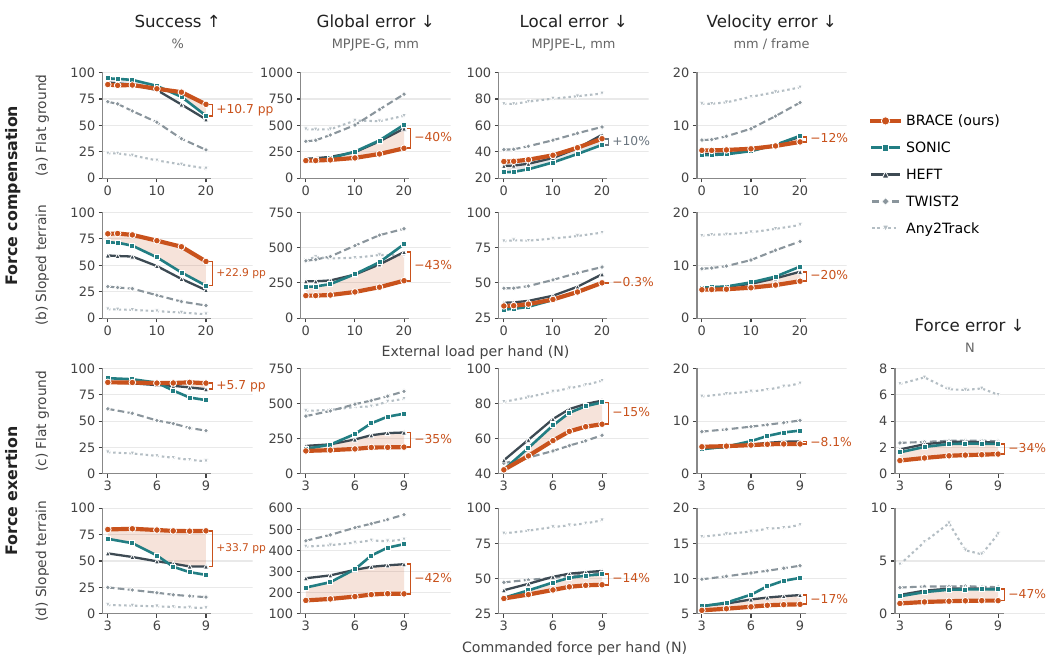}
    \caption{\textbf{Force compensation on (a) flat ground and (b) sloped terrain; force exertion on (c) flat ground and (d) sloped terrain.} Columns are success, global error (\texttt{MPJPE-G}), local error (\texttt{MPJPE-L}), velocity error, and force error (exertion only). Each point is the equal-weight mean over the PHUMA and Bones-seed test splits. The horizontal axis is the sustained load per hand (compensation, 0--20~N) or the commanded force per hand (exertion, 3--9~N). Shading marks the gap between \shortname and the better of SONIC and HEFT (orange where \shortname leads, gray where it trails); brackets give this margin at the heaviest command, in percentage points (pp) of success or as relative change in error.}
    \vspace{-6mm}
    \label{fig:quant_res}
\end{figure*}

We evaluate the whole-body tracking, force compensation, and force exertion capabilities of \shortname on flat ground and sloped terrain. Fig.~\ref{fig:quant_res} summarizes all four settings.

\subsection{Evaluation Setup}
\textbf{Baselines.} We compare with the state of the art whole motion trackers SONIC~\cite{sonic}, TWIST2~\cite{twist2}, and Any2Track~\cite{any2track}, and the payload-oriented HEFT~\cite{heft}. Force-commanding controllers~\cite{unifp,facet} are not included: neither accepts a whole-body reference trajectory, so neither admits this protocol without redefining its command interface. Evaluations use parallel GPU simulation in Isaac Lab~\cite{isaaclab} on the 29-DoF G1, retaining each baseline's native observations, action decoding, and joint gains.
All \shortname results use the deployed student, which observes neither the simulator height map nor the applied compensation wrench.

\textbf{Datasets.} We use motions from the PHUMA dataset~\cite{phuma} and a held-out test split of Bones-seed~\cite{bones_seed_2026}, built from motions that are absent from our training set. The combined test split is identical for every method.

\textbf{Protocols.} \emph{Force compensation:} with the motion reference unchanged, each hand receives a constant load of 0--20~N whose direction is sampled uniformly on a sphere and held fixed for the episode. \emph{Force exertion:} the robot receives a force command of 3--9~N per hand in the torso-heading frame and tracks the force-compensated reference, which is also what is available during real-robot deployment; a moving-reference spring-damper interaction applies the reaction in the simulator, and pose-feasibility limits can reduce the effective target. \emph{Terrain:} slopes with nominal grades of 4.6--17.7$^\circ$ (realized grade varies along the trajectory). In both terrain settings the policy receives only the flat-ground reference and must handle the slope blind from proprioception; motions are scored against terrain-conforming references.

\textbf{Metrics.} \texttt{MPJPE-G} measures the average distance between the robot's body positions and their reference positions in world coordinates. \texttt{MPJPE-L} removes root translation to measure how closely the robot matches the reference pose. A robot can therefore have low local error while drifting away from its intended position. Reported tracking errors average all recorded frames up to the first height/orientation violation.

\emph{Success criteria.} Both settings use the same completion check: throughout the full motion, the heights of the torso, wrists, and ankles must each stay within 0.25~m of their reference heights, and torso-orientation error must stay within 1~rad. The torso is the robot's tracking anchor. Exceeding either limit at any frame makes the motion unsuccessful. For motions that pass the following position requirement must be met: (a) \textit{Flat ground}: The torso's horizontal distance from its reference position, averaged over the entire motion, must be at most 0.5~m. This checks whether the robot drifts away across the ground; (b) \textit{Terrain}: We use a 3-D metric: the distance between each body and its reference position, averaged over all bodies and the entire motion (\texttt{MPJPE-G}), must be at most 0.5~m. The reference is adjusted to the terrain and includes the force brace during exertion.

In each setting, success is the percentage of all tested motions that pass both the completion check and the corresponding position check. The same thresholds apply to every basline.

\emph{Force error.} For each completed exertion motion, we compare the average measured force with the average target force after feasibility limits are applied.

\subsection{Force Compensation}

\subsubsection{Flat Ground Evaluations}\label{sec:force_comp_flat}
Fig.~\ref{fig:quant_res}(a) it can be observed without load, the performance of all baselines are comparable, while SONIC~\cite{sonic} maintaining the lower local pose error, which is expected for a tracker specialized for imitation. However,
as the load grows, every baseline degrades steadily while \shortname degrades gently, and the curves cross by 15~N.
At the highest tested compensation load, 20~N per hand, \shortname leads the best baseline by 10.7 percentage points in success.
The global-error panel explains where the baselines fail. 
Their \texttt{MPJPE-G} more than doubles over the sweep, whereas \shortname's rises far less and ends 40\% lower than the best baseline, including HEFT~\cite{heft}, which is trained for payload compensation. Local error stays within 10\% of SONIC's throughout, so the baselines are not losing pose fidelity; they are being pushed off their intended position. \shortname resists this drift without observing the load, which we attribute to the wrench-compensation teacher and its distillation into the student.

\subsubsection{Terrain Evaluations}\label{sec:force_comp_terr}
On slopes (Fig.~\ref{fig:quant_res}(b)) the ordering no longer depends on load. \shortname leads at every load, already without any hand load, and the gap widens to 22.9  percentage points at 20~N, twice the flat-ground margin. Global error tells the same story: the baselines' \texttt{MPJPE-G} rises steeply along the sweep while \shortname's stays close to its flat-ground level, ending 43\% below the best baseline. Local error is indistinguishable between \shortname and SONIC.
All methods receive the same flat-ground reference and no terrain information, so the difference lies in how each policy reconciles that reference with the slope it feels. The baselines track the flat reference literally and slide away from the target, whereas \shortname's terrain-conformed training lets it hold position on the incline while also resisting the load.

\subsection{Force Exertion}

\subsubsection{Flat Ground Evaluations}\label{sec:force_exert_flat}
In this experiment all the models are fed the flat terrain force compensated reference trajectory $s_{g}^{e}$ and the motion trackers are run. 
In the simulator a virtual spring mass damper acts on the robot in motion, and provides the necessary reaction force to simulate the real world interaction.
The overall results can be observed in, Fig.~\ref{fig:quant_res}(c) shows that the baselines can exert small forces but not large ones. At the 3~N command, SONIC and HEFT succeed on most motions, and their force errors are within about 1~N of ours. As the command increases, SONIC's success drops by more than 20 points and its global error more than doubles, while HEFT declines more gradually. \shortname's success and global error are nearly flat across the entire 3--9~N range; At 9 N, BRACE exceeds HEFT’s success rate by 5.7 percentage points and achieves 35\% lower global error, suggesting better resistance to positional drift during force exertion. \shortname also achieves the lowest force error at every command, with a 34\% reduction relative to the closest baseline at 9 N.

\subsubsection{Terrain Evaluations}\label{sec:force_exert_terr}
The terrain results in Fig.~\ref{fig:quant_res}(d) combine both demands: exerting a commanded force from diverse poses while handling an unobserved slope. \shortname's success is flat across the entire command range and matches its unloaded terrain tracking rate, while SONIC loses nearly 30 points from 3 to 9~N and HEFT starts lower and declines further. The resulting margin at 9~N is  33.7  percentage points, the largest of all four settings, with 42\% lower global error and 47\% lower force error than the closest baseline.
The four panels are consistent: the harder the setting, the larger \shortname's advantage, and the advantage comes from retaining global position and force accuracy rather than from tighter pose imitation.
These comparisons establish tracking performance under load against motion and payload trackers; relative force accuracy against force-commanding controllers remains untested.

\subsection{Torque Analysis}
\begin{figure}[t]
    \centering
    \includegraphics[width=\columnwidth]{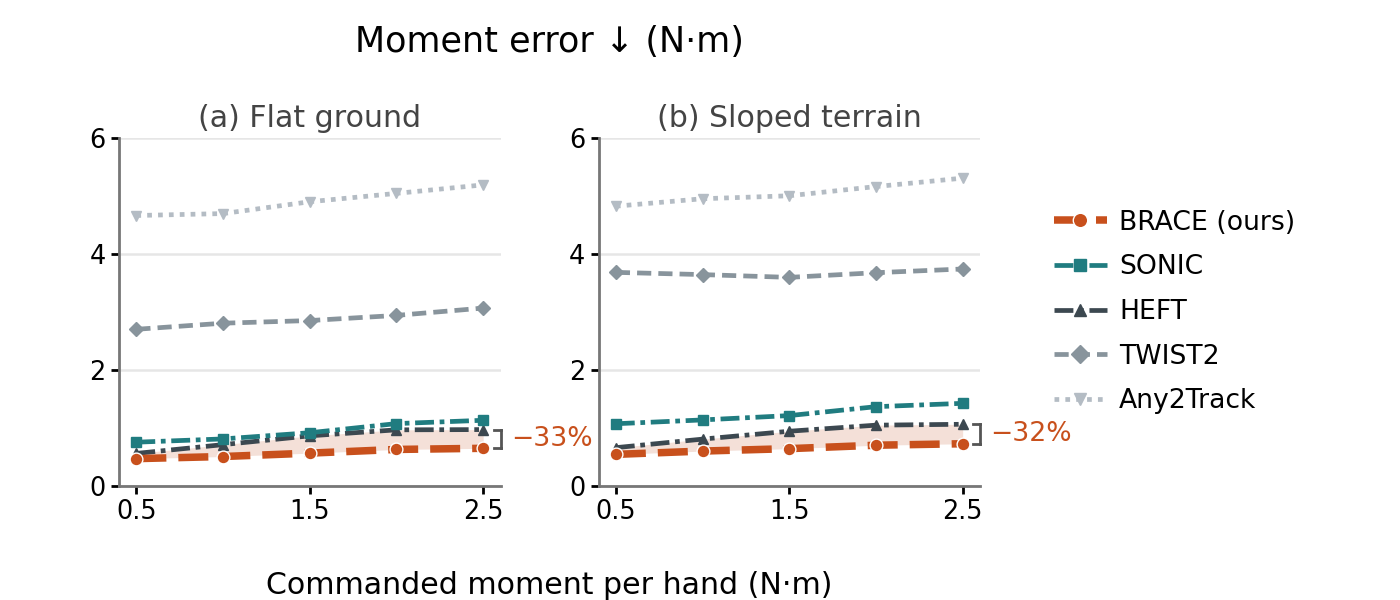}
    \caption{\footnotesize\textbf{Torque exertion error} on flat and sloped terrain.}
    \label{fig:torque}
    \vspace{-3mm}
\end{figure}

Various tasks in real world require torque (or complete wrench) exertion, for example as shown in Fig.~\ref{teaser}, to rotate the winch of a forklift a torque control is necessary.  
Fig.~\ref{fig:torque} reports moment error for commanded torques of 0.5--2.5\,N$\cdot$m on flat and sloped terrain. \shortname{} has the lowest error at every command, 33\% below the closest baseline on flat ground and 32\% on the slope. This demonstrates the ability of \shortname to accurately exert a commanded wrench. 

\begin{figure}[t]
    \centering
    \includegraphics[width=\columnwidth]{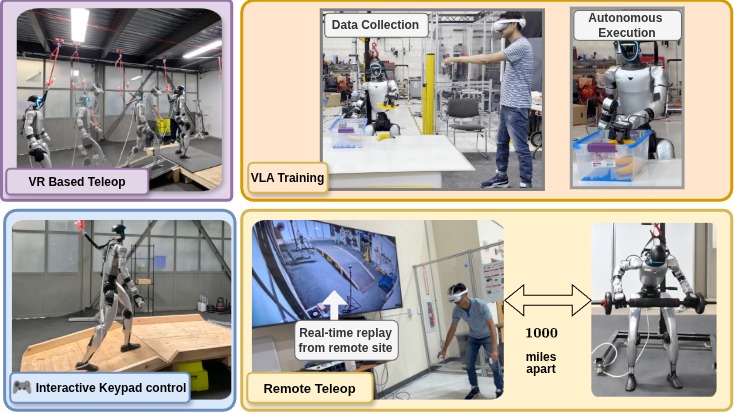}
    \caption{\footnotesize\textbf{Applications of \shortname.} Remote \& VR Teleoperation interfaces, data collection for autonomous VLA autonomous execution}
    \vspace{-6mm}
    \label{fig:applications}
\end{figure}

\subsection{Applications}\label{sec:applications}
\shortname{} supports various applications as shown in Fig.~\ref{fig:applications}. \textbf{VLA integration.} Using 362 teleoperated G1 demonstrations of transferring three bananas from a table into a container, we fine-tuned GR00T~N1.7 3B~\cite{grootn17,grootn1} to predict whole-body and Dex3 hand commands from vision, proprioception, and a fixed instruction. The policy executes autonomously on hardware. \textbf{Remote teleoperation.} We demonstrate teleoperation across sites approximately 1,000 miles apart, with \shortname{} adapting operator references to the robot's terrain. \textbf{VR teleoperation.} PICO controller poses stream through XRoboToolkit~\cite{zhao2025xrobotoolkit} and retargeted to the G1 using GMR~\cite{gmr}. \shortname{} tracks these references under varying terrain and loads. \textbf{Interactive modes.} Gamepad root-velocity commands drive MotionBricks~\cite{motion_bricks} to generate whole-body G1 trajectories given root velocity commands from gamepads \shortname{} tracks both reference sources on hardware.